\documentclass[conference]{IEEEtran}
\usepackage{graphicx}
\usepackage{booktabs}
\usepackage[table]{xcolor}
\usepackage{amsmath}
\usepackage{bm}
\usepackage{amssymb}
\usepackage[shortcuts,acronym]{glossaries}
\newacronym{isac}{ISAC}{integrated sensing and communication}
\newacronym{fid}{FID}{Fisher information density}
\newacronym{fim}{FIM}{Fisher information matrix}
\newacronym{plr}{PLR}{privacy leakage ratio}
\newacronym{ael}{AEL}{average exposure length}
\newacronym{mel}{MEL}{mean maximum exposure length}
\newacronym{crb}{CRB}{Cramer-Rao bound}
\newacronym{slam}{SLAM}{simultaneous localization and mapping}
\newacronym{ucy}{UCY}{University of Cyprus}
\newacronym{v2i}{V2I}{vehicle-to-infrastructure}
\newacronym{uav}{UAV}{uncrewed aerial vehicle}
\newacronym{snr}{SNR}{signal-to-noise ratio}
\newacronym{gdpr}{GDPR}{General Data Protection Regulation}
\newacronym{mae}{MAE}{mean absolute error}

\begin{document}

\title{GDPR-Aware Trajectory Sharing for ISAC-Assisted Robot Navigation: A Case Study on FID-Constrained Collision Prediction}

\author{
    \IEEEauthorblockN{Zexin~Fang\IEEEauthorrefmark{1},~Bin~Han\IEEEauthorrefmark{1},~Donglin~Wang\IEEEauthorrefmark{1},~Fengchen~Pei\IEEEauthorrefmark{2} and~Hans~D.~Schotten\IEEEauthorrefmark{1}\IEEEauthorrefmark{3}}
    \IEEEauthorblockA{
  \IEEEauthorrefmark{1}{RPTU University Kaiserslautern-Landau, Germany}; \IEEEauthorrefmark{2}{Technical University of Darmstadt, Germany}\\ 
  \IEEEauthorrefmark{3}{German Research Center for Artificial Intelligence (DFKI), Germany.}}
}

\maketitle

\begin{abstract}
\Gls{isac} enables intelligent wireless infrastructure but raises growing
regulatory concern as fine-grained personal trajectory histories become a
byproduct of sensing. \Gls{gdpr} Articles~5(1)(c) and~5(1)(f) require that
personal data be limited to what is necessary and protected through appropriate
technical measures against unauthorised reconstruction. This paper addresses
both requirements through a \gls{fid}-constrained trajectory sharing scheme for
robot collision avoidance, where sensing estimates are perturbed according to
local information content before sharing. Experiments on real pedestrian traces 
show that \gls{fid}-controlled sharing achieves a strictly
better privacy-utility tradeoff than fixed-error perturbation: at matched
missed-conflict rates, reconstruction leakage and sustained exposure lengths
are consistently lower, establishing information-aware perturbation as a
principled technical measure aligned with \gls{gdpr} data minimisation and
integrity requirements.
\end{abstract}

\begin{IEEEkeywords}
GDPR, ISAC, 6G, privacy.
\end{IEEEkeywords}

\glsresetall

\section{Introduction}
\label{sec:introduction}

\Gls{isac} is reshaping wireless infrastructure into a shared perception layer,
where the same radio measurements that support communication also enable radio
mapping, channel and beam prediction, cooperative perception, and
mobility-aware resource allocation
\cite{zhou2024temporal,yang2025cooperative,li2026attention,li2026recent}.
For mobile robots and autonomous agents, this creates a compelling operational
advantage: rather than reacting to an imminent collision, a robot can access
shared trajectory histories of surrounding agents and anticipate path conflicts
before they materialize \cite{liu2026goalo,ye2024isac,selvaraj2024edge}.
Trajectory sharing thus becomes a useful sensing primitive, not merely a
communication-side optimization.

This capability, however, introduces a privacy problem that is more immediate
and pervasive than the high-resolution risks typically discussed in
\gls{isac} literature. Prior work has focused on sensitive inference targets
such as human activity recognition, vital-sign estimation, and physiological
monitoring \cite{li2022integrated,gunlu2026isacprivacy}. These are serious
concerns for future deployments, but the most direct risk in mobility-oriented
\gls{isac} is already present: repeated radio sensing produces accurate
personal trajectory histories that expose not only position, but speed,
heading, stopping behavior, route choice, and recurring mobility patterns.
Recent event-level \gls{isac} work further argues that sensing is shifting
from isolated target snapshots toward behavioral semantics and intent
prediction \cite{liu2026event}, meaning that shared trajectory data can support
behavior inference even without any biometric classifier.

The tension is therefore not only a generic privacy-utility tradeoff, but a
\gls{gdpr}-style data minimisation question: a robot navigating among pedestrians
needs enough trajectory information to predict collision risk, but not a precise
reconstruction of each person's motion history. Under data minimisation
principles, personal data should not be processed, stored, or shared beyond
what the stated purpose strictly requires. The shared trajectory should
therefore be no more accurate, dense, or continuous than the safety task
demands, and any precision beyond that threshold constitutes an unjustified
exposure. Crucially, this remains a privacy concern even when track labels are
randomised or rotated across sessions. A trajectory released at high spatial and
temporal resolution carries enough internal consistency, in terms of speed
profile, turning behavior, and stopping patterns, for an observer to re-identify
and link records across time through correlation or similarity analysis, without
any stable identifier. The privacy risk therefore does not hinge on whether the
same label persists; it hinges on whether the shared data is accurate and
consistent enough to make such linkage feasible.
 
We study this tension through a concrete case: predictive collision detection for a robot moving among 
pedestrians. This task captures the core \gls{isac} advantage, namely that the robot benefits from shared
 trajectory histories before its own sensors would detect an impending conflict, and it concentrates the privacy 
 risk precisely where useful data and sensitive data coincide. \Gls{isac}-assisted \gls{v2i} systems have shown 
 that sensing supports collision avoidance beyond localization \cite{ye2024isac}, cooperative sensing has been 
 applied to obstacle estimation in \gls{uav} formations \cite{wang2025cooperative}, and edge-assisted studies 
 rely on trajectory prediction and uncertainty-aware conflict forecasting to trigger braking before imminent 
 collisions \cite{selvaraj2024edge}. The collision-prediction task used here is therefore not an artificial 
 add-on to \gls{slam} or mapping; it captures the forecasting layer that connects sensing, shared trajectory 
 data, and safe robot control in dynamic scenes.

Our approach builds on the \gls{fid}-constrained sharing framework from prior work. Rather than applying uniform 
perturbation, we scale added noise to each trajectory segment according to its local information content: highly 
informative segments receive stronger perturbation, while already uncertain segments are changed less. This avoids 
the failure modes of fixed-noise schemes, which are simultaneously too weak where sensing is most accurate and 
unnecessarily destructive where estimates are already poor. The goal is not to introduce a new privacy metric, 
but to validate this existing mechanism in a realistic setting, using real pedestrian trajectory traces. Results 
show that \gls{fid}-controlled perturbation achieves a better privacy-utility tradeoff than fixed-noise schemes: 
at the same collision prediction failure rate, significantly stronger privacy protection is retained.

The rest of this paper is organized as follows. Section~\ref{sec:fid_sharing}
summarizes the \gls{fim}/\gls{fid}-constrained sharing model.
Section~\ref{sec:swarm_dataset} describes the scene and dataset.
Section~\ref{subsec:swarm_validation} defines the collision-prediction task and
privacy descriptors. Section~\ref{sec:results} presents the results, and
Section~\ref{sec:conclusion} concludes the paper.

\section{\gls{fim}/\gls{fid}-Constrained Trajectory Sharing}
\label{sec:fid_sharing}
\label{subsec:privacy_inference_bridge}

 Let $\bm{m}_i(t_k)$ denote the ground-truth
trajectory sample of agent $i$, $\hat{\bm{m}}_i(t_k)$ the raw \gls{isac} sensing
estimate, and $\tilde{\bm{m}}_i(t_k)$ the trajectory sample shared with the
robot or a network entity. As in the previous \gls{fid}-sharing model
\cite{sensing2025fang}, the sensing
uncertainty is characterized through the \gls{crb}
\cite{crb2024ren,fang2026balancing}, or equivalently through the
Fisher information obtained at sensing update $k$:
\begin{equation}
    \mathcal{I}_i(k)
    =
    \beta_i(k)m_{s,i}(k)\mathbf{a}_i^H(k)
    \mathbf{Q}_{s,i}(k)\mathbf{a}_i(k),
    \label{eq:swarm_fi}
\end{equation}
where $\beta_i(k)$ captures propagation and target-dependent gain,
$m_{s,i}(k)$ is the sensing resource allocation, $\mathbf{a}_i(k)$ is the
steering vector, and $\mathbf{Q}_{s,i}(k)$ is the sensing transmit covariance.
The local information density used for privacy control follows the discrete
\gls{fid} definition in the system model,
\begin{equation}
    \mathcal{J}_i(t)
    =
    \frac{\mathcal{I}_i(k)}{t_k-t_{k-1}},
    \quad t\in(t_{k-1},t_k].
    \label{eq:swarm_fid}
\end{equation}

The shared trajectory is then generated from the raw sensing estimate using
the \gls{fid}-controlled perturbation mechanism from the previous privacy
defense model \cite{sensing2025fang}:
\begin{align}
    \tilde{\bm{m}}_i(t_k)
    &=
    \hat{\bm{m}}_i(t_k) + \Delta \bm{e}_i(t_k), \nonumber\\
    \Delta \bm{e}_i(t_k)
    &\sim
    \mathcal{N}(\bm{0},(\Delta\sigma(t_k))^2).
    \label{eq:swarm_shared}
\end{align}
The additional standard deviation is determined by the local \gls{fid}
threshold ratio
\begin{equation}
    \rho_i(k)=\frac{\mathcal{J}_i(t_k)}{\eta},
    \label{eq:fid_ratio}
\end{equation}
where $\eta$ is the information-density constraint. The error-control rule used
in the simulation is
\begin{equation}
    \Delta\sigma_i(t_k)=
    \begin{cases}
        0, & \rho_i(k)\leq 1,\\
        \alpha\left(\beta-\exp[-(\rho_i(k)-1)]\right),
        & \rho_i(k)>1,
    \end{cases}
    \label{eq:fid_error_control}
\end{equation}
and the shared sensing standard deviation becomes
\begin{equation}
    \sigma^{\mathrm{sh}}_i(t_k)
    =
    \sigma^{\mathrm{raw}}_i(t_k)+\Delta\sigma_i(t_k).
    \label{eq:shared_sigma}
\end{equation}
Thus, high-\gls{fim} segments, which are more accurately sensed and more
informative, receive additional perturbation before sharing. In the reported
experiments, $\alpha=0.5$ and $\beta=1.5$.

The three privacy descriptors are used because they indicate trajectory-level
linkability. We do not claim that \gls{plr}, \gls{ael}, and \gls{mel} directly
measure civil-identity re-identification. Instead, they quantify whether a
pseudonymous track remains continuously reconstructable with enough precision
to follow the same agent over time. For a trajectory with $K$ shared samples,
define the leakage indicator
\begin{equation}
    z_i(k)
    =
    \mathbf{1}
    \left[
    \left\|\tilde{\bm{m}}_i(t_k)-\bm{m}_i(t_k)\right\| < \epsilon
    \right].
    \label{eq:leakage_indicator}
\end{equation}
The privacy leakage ratio is
\begin{equation}
    \mathrm{PLR}_i
    =
    \frac{1}{K}\sum_{k=1}^{K}z_i(k).
    \label{eq:plr_definition}
\end{equation}
Consecutive nonzero entries of $z_i(k)$ form leakage segments. We report
\gls{ael} as the typical continuous duration of a leaked segment and \gls{mel}
as the longest continuous exposure within each trajectory before averaging
across Monte Carlo trials. Thus, the metrics capture both pointwise leakage and
the continuous trajectory evidence needed to infer route, stops, velocity, and
motion habits.

\section{Scene and Dataset Construction}
\label{sec:swarm_dataset}

In this paper, each Monte Carlo scene contains one robot and four moving agents. The robot is
the decision maker: it receives the shared trajectory histories of nearby agents
and checks whether its nominal path will conflict with their predicted motion.
The four agents represent surrounding pedestrians whose motion is not controlled
by the robot.

Agent motion is drawn from the \gls{ucy} Students trajectory set distributed
through OpenTraj. We use the raw Students03 observation file rather than
synthetic straight-line tracks. Each pedestrian track is grouped by track ID,
converted to metric coordinates, and discarded if it is too short for the
history/future split. The remaining tracks are cubically interpolated and
sampled using the same adaptive procedure as the prior sensing pipeline. In the
reported sensitivity setting, the update density is doubled to $8$--$16$
samples/s while preserving the original temporal scale, yielding an average of
$366$ shared history points per agent at a mean sampling interval of $0.083$\,s.

For each scene, four pedestrian traces are selected cyclically from the
available \gls{ucy} tracks and each is split into a known history and a hidden
future at a $70\%$ ratio. The last history sample serves as the agent's current
position; the future segment determines the ground-truth conflict label. The robot moves at $1.4$\,m/s, giving a human-like traversal speed over a short
horizon.

To obtain both positive and challenging negative cases, each pedestrian trace
is translated spatially while preserving its original shape, timing, and
velocity profile. In conflict scenes, the first agent is repositioned so that
its future path crosses the robot's nominal trajectory at a randomly chosen
time step. In non-conflict scenes, the same agent is repositioned so that its
future path passes just outside the $0.9$\,m safety radius, forming a near
miss rather than a collision. The remaining three agents are placed around the
robot as background distractors. Because the underlying motion always comes
from real pedestrian recordings, the constructed scenes remain behaviorally
realistic while guaranteeing that the evaluation includes nontrivial cases on
both sides of the conflict boundary.

\section{Predictive Navigation Validation}
\label{subsec:swarm_validation}

We evaluate whether \gls{fid}-constrained shared trajectories retain sufficient
motion information for predictive collision detection. One robot follows a short
nominal path while observing four surrounding agents. For each agent, the robot
receives a shared trajectory history and estimates future motion from the most
recent samples, then checks whether the predicted path conflicts with its own
nominal route over a $12$-step horizon.

Let $\tilde{\bm{m}}_i(t_k)=[\tilde{x}_i(t_k),\tilde{y}_i(t_k)]^T$ denote the
most recent shared position of agent $i$. The robot fits two first-order
polynomials by least squares over the $q=16$ most recent shared samples,
\begin{equation}
\tilde{x}_i(t)=a_{x,i}t+b_{x,i},\quad
\tilde{y}_i(t)=a_{y,i}t+b_{y,i},
\label{eq:linear_fit}
\end{equation}
giving the estimated velocity
\begin{equation}
\tilde{\bm{v}}_i=[a_{x,i},\,a_{y,i}]^T.
\label{eq:velocity_estimate}
\end{equation}
Future timestamps are generated from the mean sampling interval of the recent
window. For horizon index $h\in\{0,\ldots,H-1\}$ with $H=12$,
\begin{equation}
\hat{t}_{k+h}=t_k+h\bar{\Delta t},\quad
\bar{\Delta t}=\frac{1}{q-1}\sum_{\ell=k-q+2}^{k}(t_\ell-t_{\ell-1}),
\label{eq:future_time_grid}
\end{equation}
and the predicted future position is
\begin{equation}
\hat{\bm{m}}_i(\hat{t}_{k+h})
=[a_{x,i}\hat{t}_{k+h}+b_{x,i},\;
a_{y,i}\hat{t}_{k+h}+b_{y,i}]^T.
\label{eq:agent_prediction}
\end{equation}
The robot path over the same $H$ steps is
\begin{equation}
\bm{r}_h=\bm{r}_0+
\min\!\left(v_{\mathrm{r}}h\bar{\Delta t},\,
\|\bm{d}-\bm{r}_0\|\right)
\frac{\bm{d}-\bm{r}_0}{\|\bm{d}-\bm{r}_0\|},
\label{eq:robot_nominal_path}
\end{equation}
where $\bm{r}_0$ is the robot's current position, $\bm{d}$ is its destination,
and $v_{\mathrm{r}}=1.4$\,m/s. A predicted conflict is declared when the
minimum time-aligned separation falls below the safety radius,
\begin{equation}
\hat{c}=\mathbf{1}\!\left[
\min_{i,h}\left\|\bm{r}_h-\hat{\bm{m}}_i(\hat{t}_{k+h})\right\|
<d_{\mathrm{safe}}\right],
\label{eq:conflict_rule}
\end{equation}
with $d_{\mathrm{safe}}=0.9$\,m. The ground-truth label $c$ is obtained by the
same rule applied to the true hidden future trajectories $\bm{m}_i(t_{k+h})$.
A missed conflict occurs when $c=1$ and $\hat{c}=0$. The distance \gls{mae}
is the absolute error between the predicted and true minimum separation.
This deliberately simple predictor ensures that any degradation in conflict
detection is attributable to the quality of the shared trajectory rather than
to predictor complexity.

\begin{table}[t]
    \centering
    \caption{Simulation setup for predictive collision validation}
    \label{tab:swarm_setup}
    \scriptsize
    \setlength{\tabcolsep}{3pt}
    \renewcommand{\arraystretch}{1.08}
    \begin{tabular}{p{0.22\columnwidth}p{0.22\columnwidth}p{0.18\columnwidth}p{0.28\columnwidth}}
        \toprule
        \rowcolor{gray!20}
        \textbf{Category} & \textbf{Parameter} & \textbf{Value} & \textbf{Remark} \\
        \midrule
        \rowcolor{gray!8}
        Dataset/scene & Monte Carlo scenes & $500$ & Shared across all schemes \\
        \rowcolor{gray!8}
        & Moving agents & $4$ pedestrians + robot & Near-miss negatives included \\
        \midrule
        \rowcolor{gray!15}
        Robot task & Robot speed & $1.4$ m/s & Human-like speed \\
        \rowcolor{gray!15}
        & Prediction horizon & $H=12$ & Future conflict check \\
        \rowcolor{gray!15}
        & Velocity window & $q=16$ & Least-squares trend estimate \\
        \rowcolor{gray!15}
        & Safety radius & $d_{\mathrm{safe}}=0.9$ m & Conflict threshold \\
        \midrule
        \rowcolor{gray!8}
        Sensing/channel & Base-station position & $[5,30]^T$ & 2-D map coordinate \\
        \rowcolor{gray!8}
        & Carrier frequency & $3.5$ GHz & \\
        \rowcolor{gray!8}
        & Bandwidth & $100$ MHz & Sensing bandwidth \\
        \rowcolor{gray!8}
        & Multipath number & $N_p=4$ & Urban multipath model \\
        \rowcolor{gray!8}
        & Max. delay spread & $2\times10^{-7}$ s & Channel delay spread \\
        \rowcolor{gray!8}
        & Effective power & $30$ dBm & $+15$ dB sensitivity case \\
        \rowcolor{gray!8}
        & Noise floor & $N_0=-91$ dBm & Receiver noise floor \\
        \rowcolor{gray!8}
        & Path-loss exponent & $2.7$ & Average path-loss component \\
        \rowcolor{gray!8}
        & Beam fluctuation & $\mathcal{N}(0,2)$ dB & Beam misalignment effect \\
        \midrule
        \rowcolor{gray!15}
        Privacy control & Leakage threshold & $\epsilon=0.3$ m & Pointwise leakage rule \\
        \rowcolor{gray!15}
        & \gls{fid} thresholds & $\eta=1$--$1000$ sweep & Information constraint \\
        \rowcolor{gray!15}
        & Perturbation parameters & $\alpha=0.5$, $\beta=1.5$ & Eq.~\eqref{eq:fid_error_control} \\
        \rowcolor{gray!15}
        & Fixed-error baseline & $\Delta\sigma=0$--$1.0$ m sweep & Scheme comparison \\
        \bottomrule
    \end{tabular}
\end{table}

\section{Simulation Results}
\label{sec:results}

The simulation setup is listed in Table~\ref{tab:swarm_setup}. Results are
averaged over $500$ Monte Carlo scenes. In each scene, pedestrian traces are
interpolated to the sensing update grid, corrupted by \gls{snr}-dependent
\gls{crb} noise, and then shared either with a fixed added error or with the
\gls{fid}-controlled perturbation in Eq.~\eqref{eq:fid_error_control}. The robot
estimates local velocity from the shared history and applies
Eq.~\eqref{eq:conflict_rule}. Privacy is evaluated on the same shared samples
using \gls{plr}, \gls{ael}, and \gls{mel}. Near-miss negative scenes are included
to avoid trivial non-conflict cases.

\subsection{FID Threshold Sensitivity}

\begin{figure}[t]
    \centering
    \includegraphics[width=0.95\linewidth]{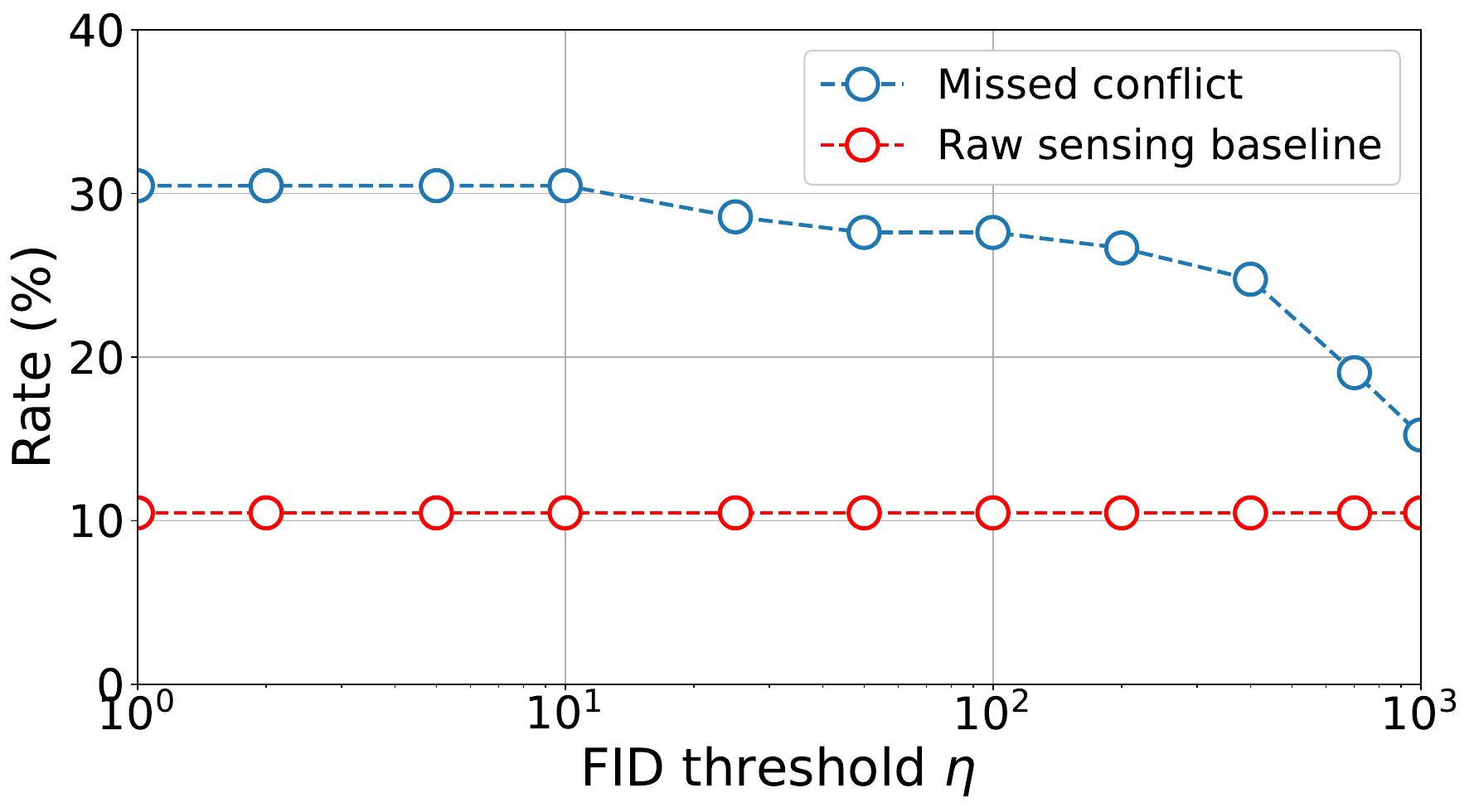}
    \vspace{0.25em}
    \includegraphics[width=0.95\linewidth]{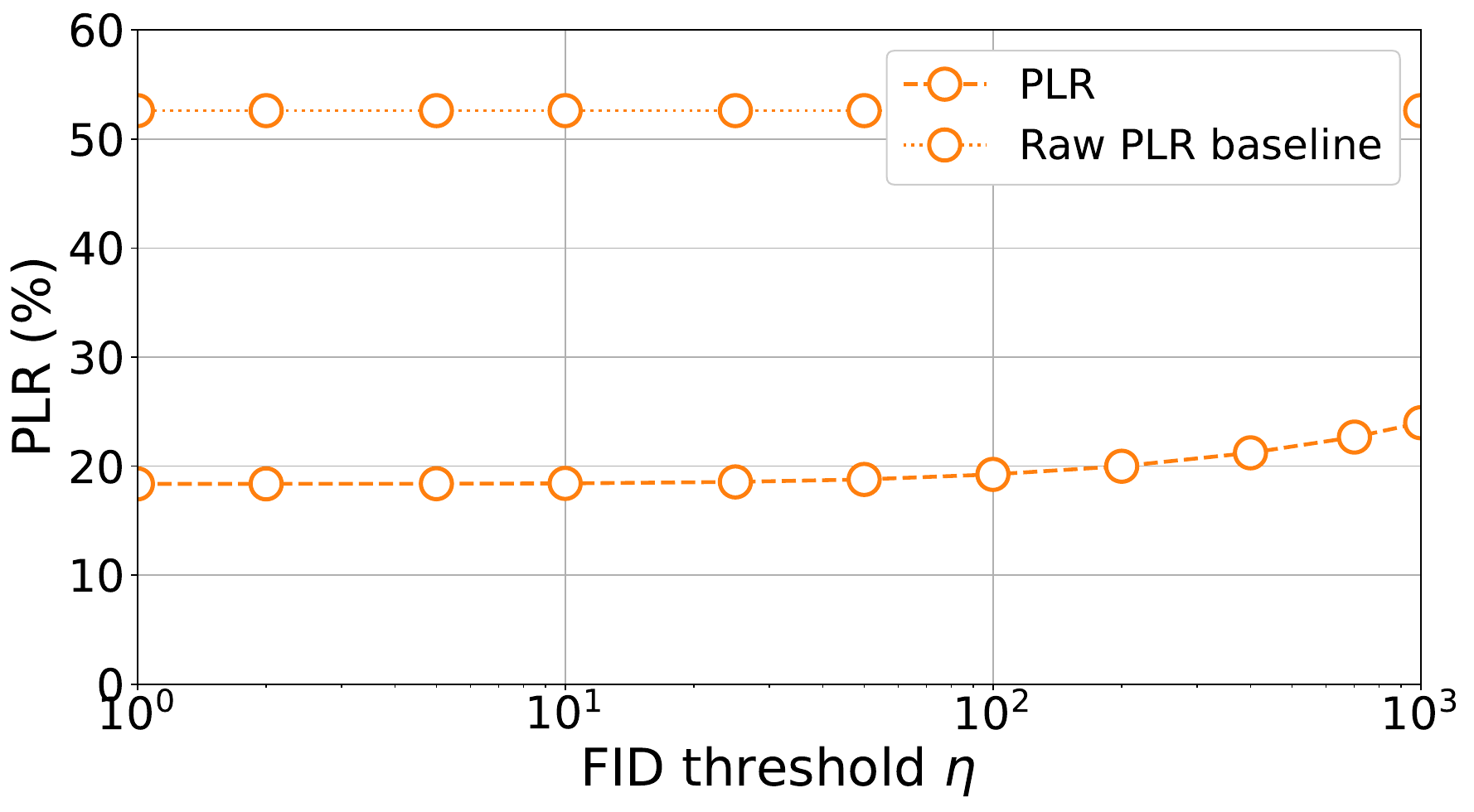}
    \vspace{0.25em}
    \includegraphics[width=0.95\linewidth]{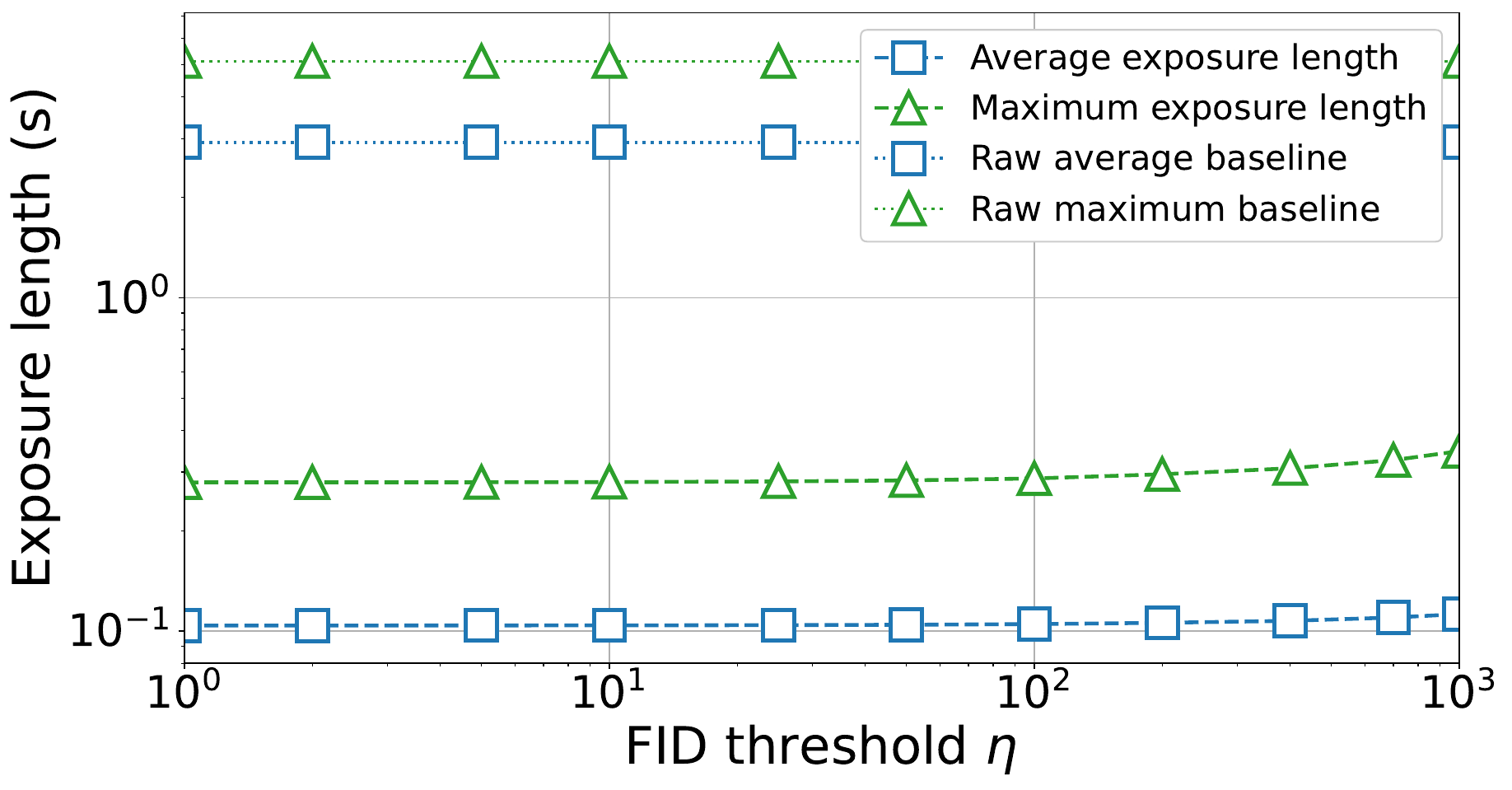}
    \caption{\gls{fid}-threshold sensitivity of missed-conflict rate,
    \gls{plr}, and exposure lengths. Dashed lines connect sampled \gls{fid}
    thresholds; dotted marker lines show raw sensing baselines.}
    \label{fig:fid_tradeoff_summary}
\end{figure}

Fig.~\ref{fig:fid_tradeoff_summary} shows the effect of varying the information
threshold $\eta$. Without privacy perturbation, the raw sensing baseline yields
a missed-conflict rate of $10.5\%$. Activating \gls{fid}-controlled sharing raises
this to approximately $27\%$, since stronger perturbation is applied precisely
where sensing is most accurate and where the velocity estimate is most
informative. Sweeping $\eta$ from $1$ to $1000$ reveals a monotone tradeoff:
the missed-conflict rate falls from $30.5\%$ at saturated protection to $15.2\%$
at $\eta=1000$, while \gls{plr} rises from $18.4\%$ to $24.0\%$ and the mean
maximum exposure length increases from $0.28$\,s to $0.35$\,s. The underlying
mechanism is straightforward: a small $\eta$ activates perturbation whenever the
local \gls{fid} exceeds the threshold, breaking the spatial continuity of the shared
track and reducing the chance that consecutive samples fall within the
reconstruction threshold $\epsilon$. This weakens linkability even when the
pseudonymous label remains available. A large $\eta$ relaxes the constraint,
allowing the shared trajectory to remain closer to the raw sensing output and
recovering more reliable short-horizon velocity trends for the robot. Even at
$\eta=1000$, however, all privacy descriptors remain well below the raw sensing
baseline (\gls{plr} $52.6\%$, average exposure $2.93$\,s, mean maximum exposure
$5.12$\,s), confirming that a single threshold jointly governs navigation utility
and trajectory leakage.

\subsection{Fixed-Noise Comparison}

\begin{figure}[t]
    \centering
    \includegraphics[width=0.95\linewidth]{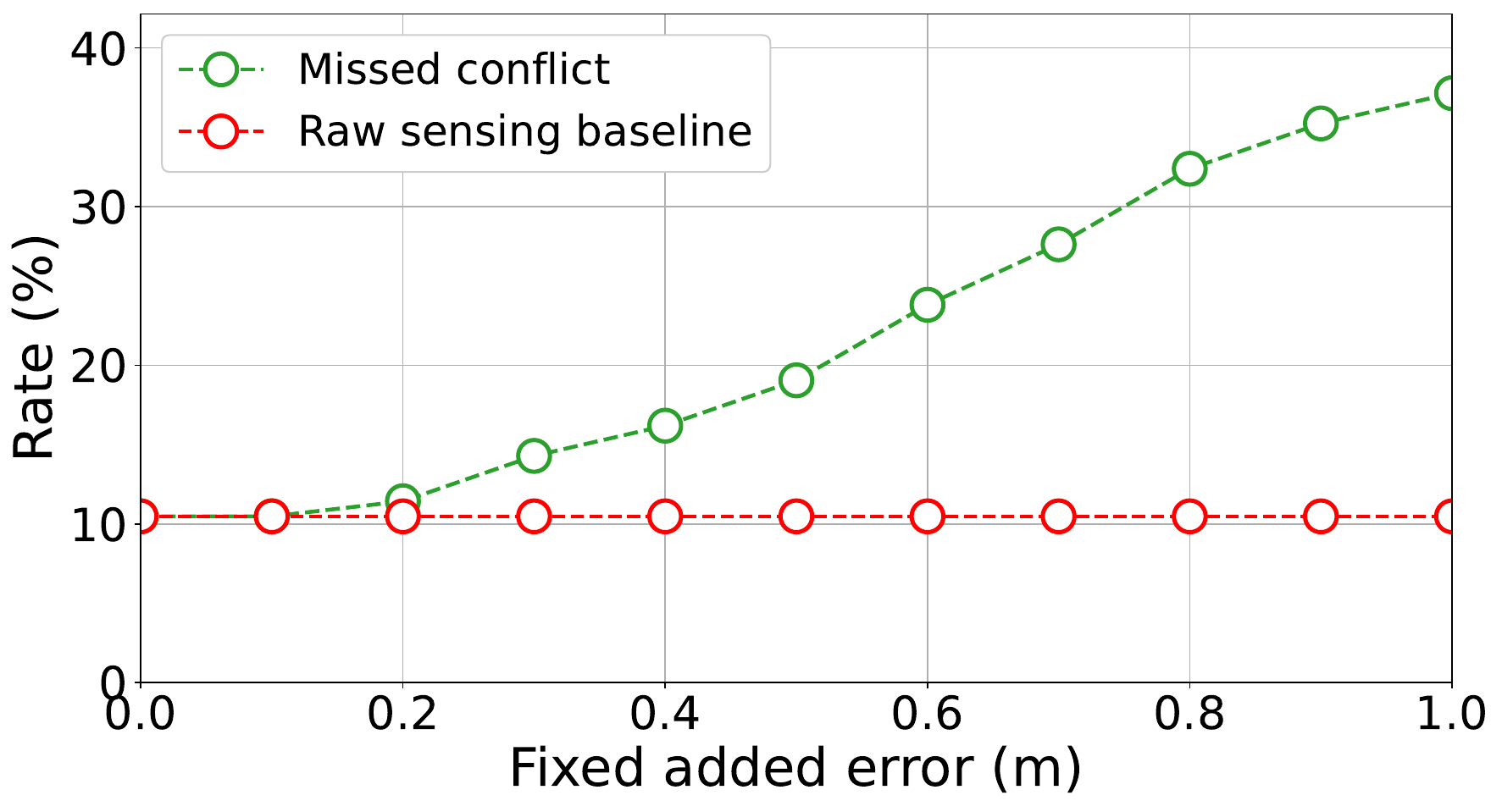}
    \vspace{0.25em}
    \includegraphics[width=0.95\linewidth]{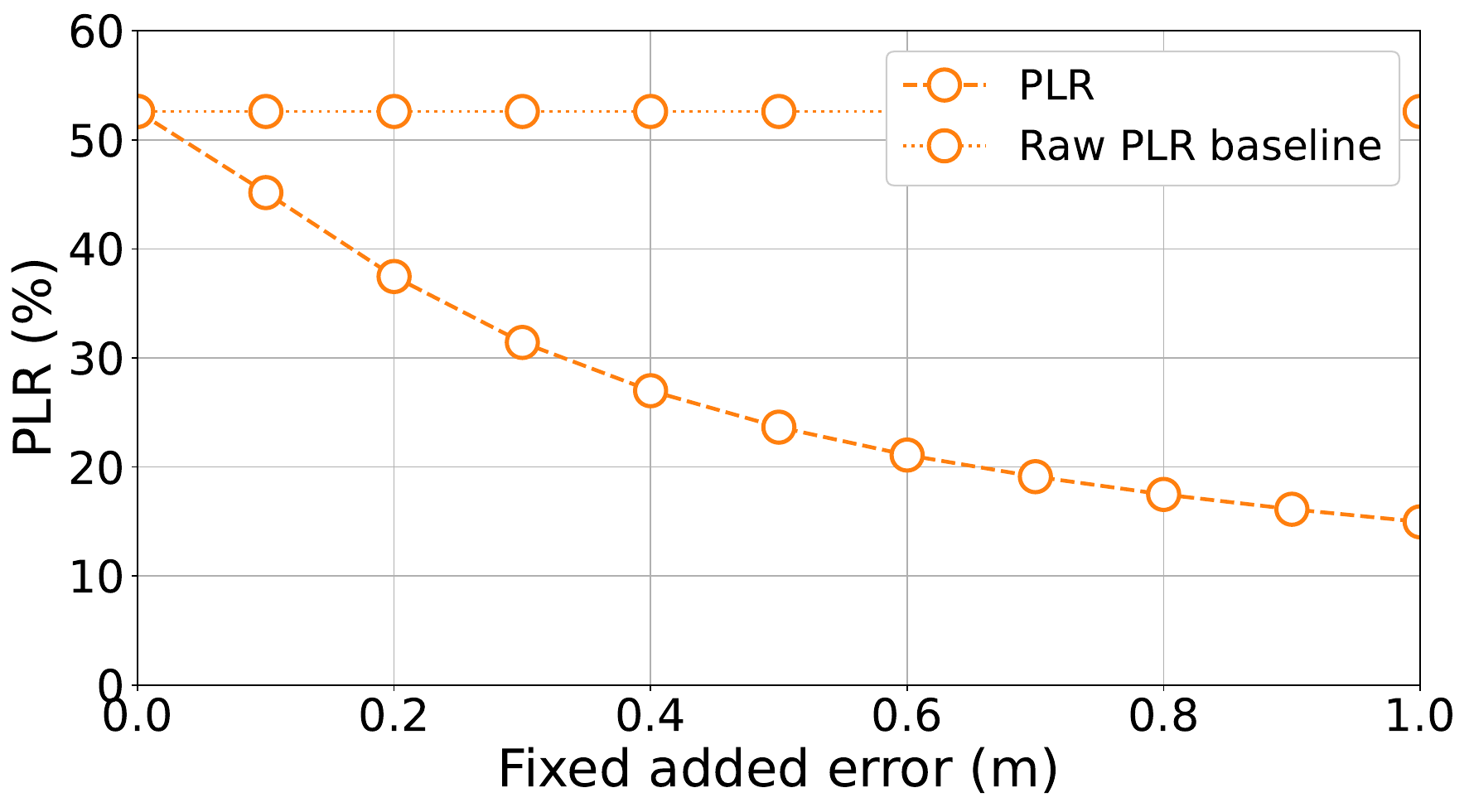}
    \vspace{0.25em}
    \includegraphics[width=0.95\linewidth]{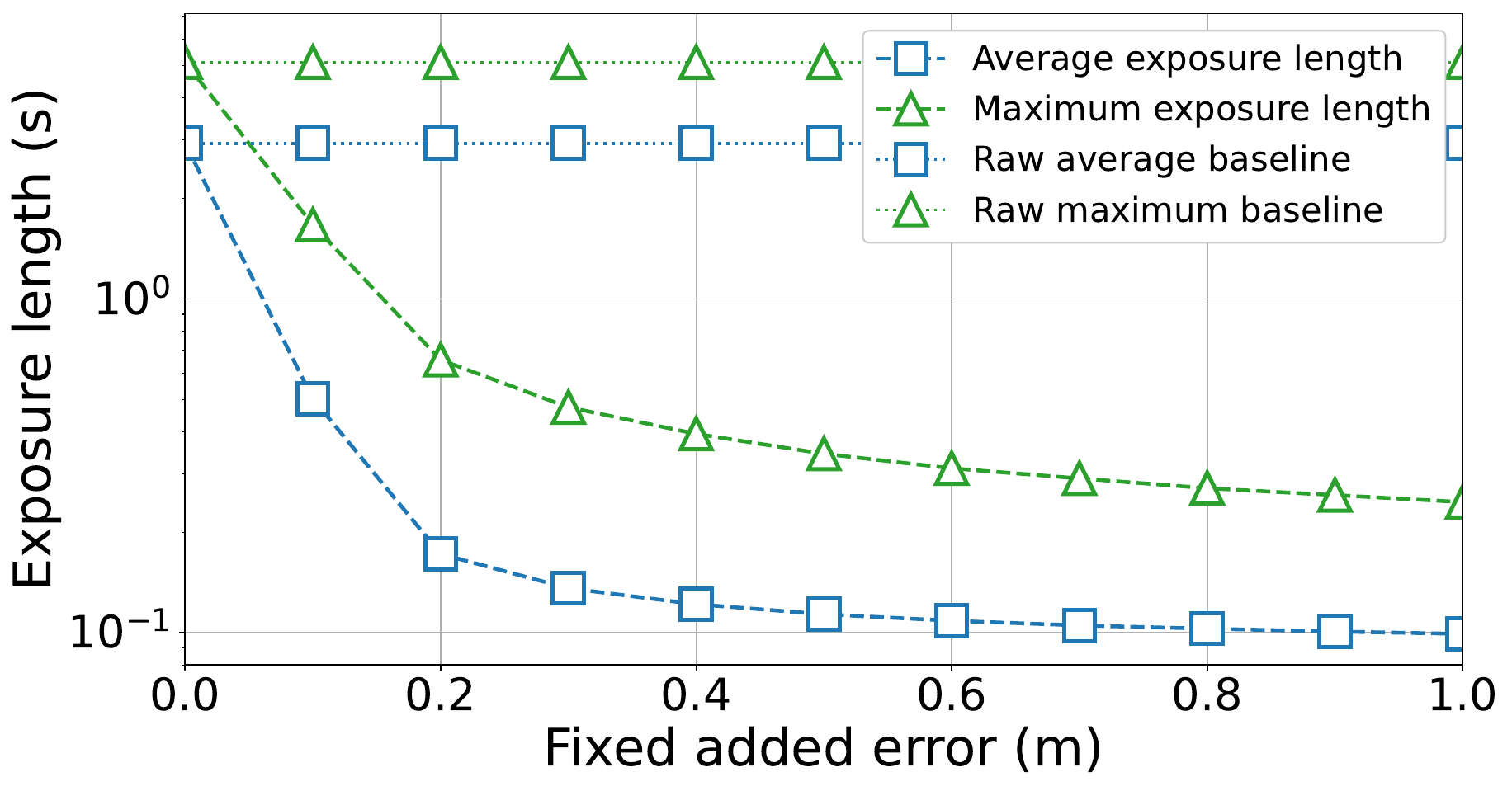}
    \caption{Fixed-error sensitivity of missed-conflict rate, \gls{plr},
    and exposure lengths under the same scene and sensing configuration.}
    \label{fig:fixed_noise_sweep}
\end{figure}

Fig.~\ref{fig:fixed_noise_sweep} gives the fixed-noise baseline. Small fixed
errors up to $0.2$\,m leave collision prediction close to the raw baseline,
while larger errors degrade it sharply: the missed-conflict rate reaches $27.6\%$
at $\Delta\sigma=0.7$\,m and $37.1\%$ at $\Delta\sigma=1.0$\,m. Privacy
improves in the opposite direction, with \gls{plr} falling from $52.6\%$ to
$15.0\%$ and mean maximum exposure length from $5.12$\,s to $0.25$\,s as the
fixed error increases. The key limitation of this scheme is that the same
additional uncertainty is applied to every shared point regardless of its local
information content. Low-information samples, which were already poorly sensed,
receive unnecessary distortion that degrades the velocity estimate without
meaningfully reducing leakage. High-information samples, where linkage risk is
greatest, may still fall within $\epsilon$ if the fixed error is too small.
\gls{fid}-controlled sharing avoids both failure modes by tying perturbation
strength to the local sensing quality.

\subsection{Operating-Point Comparison and Design Interpretation}

\begin{figure}[t]
    \centering
    \includegraphics[width=0.95\linewidth]{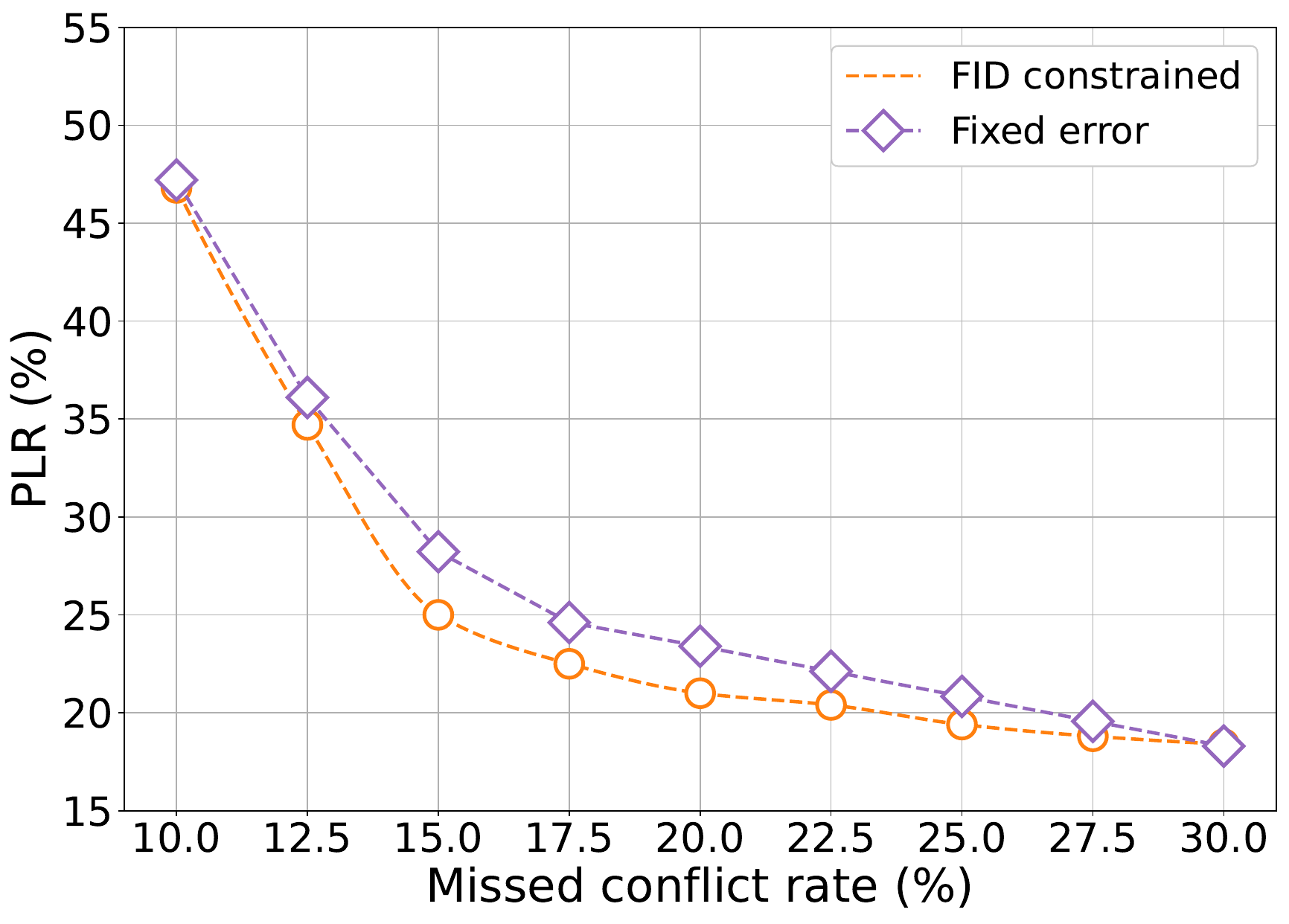}
    \vspace{0.25em}
    \includegraphics[width=0.95\linewidth]{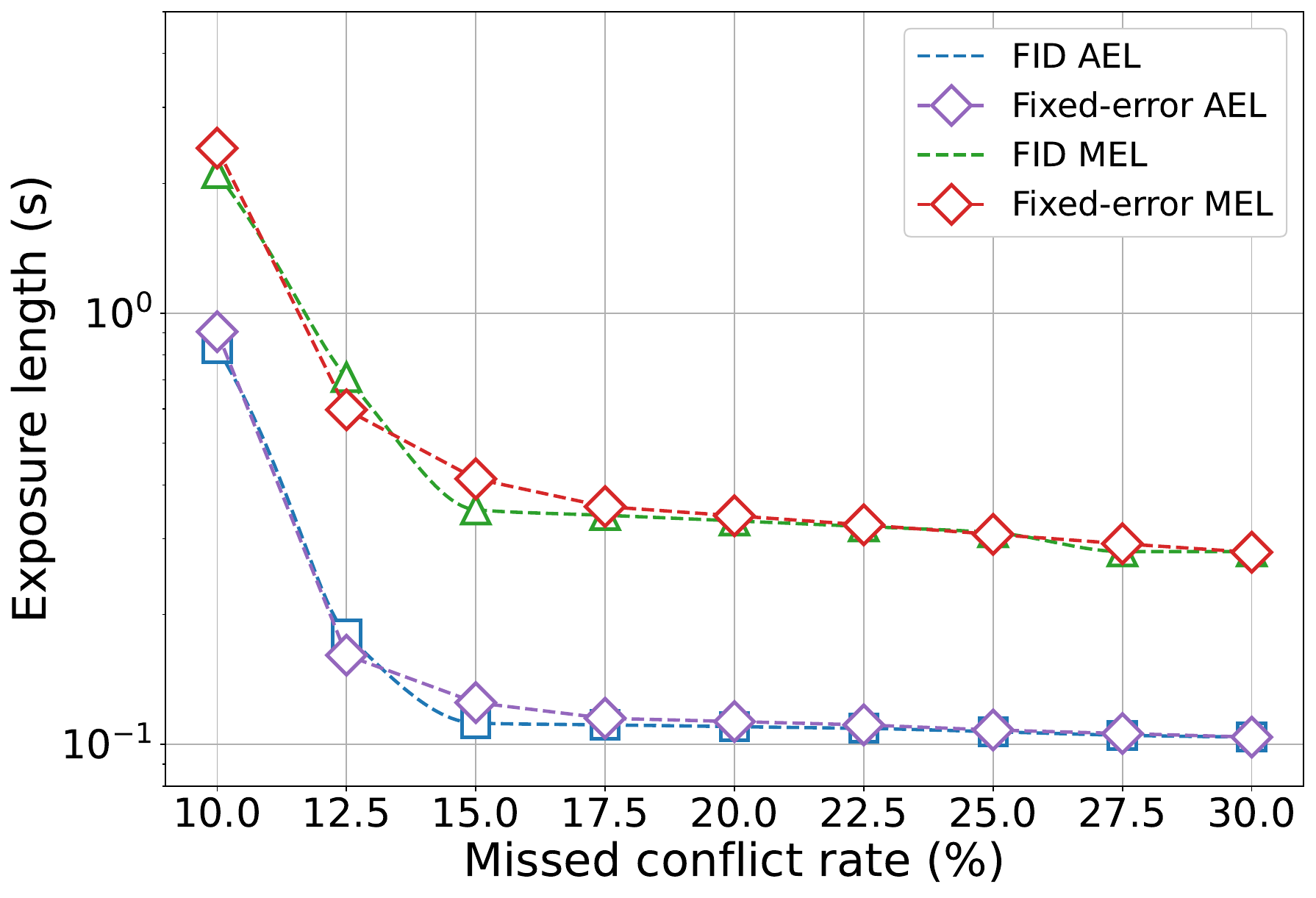}
    \caption{Operating-point comparison between \gls{fid}-controlled sharing
    and fixed-error perturbation at matched missed-conflict rates.}
    \label{fig:fid_single_point}
\end{figure}

Fig.~\ref{fig:fid_single_point} places both mechanisms on a common
privacy-utility plane by matching them at the same missed-conflict rate.
At every operating point, \gls{fid}-controlled sharing achieves lower \gls{plr}
and shorter exposure lengths than fixed-noise perturbation. This advantage
reflects where distortion is placed along the trajectory: information-aware
perturbation concentrates added noise on the most accurately sensed segments,
precisely those that would otherwise provide the sustained positional precision
needed for correlation-based re-identification. Fixed-noise perturbation, by
contrast, spreads distortion uniformly and therefore either over-protects
low-information segments or under-protects high-information ones.

This interpretation suggests a practical design workflow. The system first
selects a target missed-conflict rate compatible with the safety requirement.
Among all \gls{fid} thresholds that satisfy that target, the designer then chooses
the smallest $\eta$, which yields the strongest privacy protection at acceptable
navigation cost. The resulting \gls{plr} and exposure lengths can then be checked
directly against the applicable data minimisation requirement, for instance
whether the mean continuous exposure duration falls below a regulatory threshold.
This is qualitatively different from tuning a fixed error level, where the
privacy outcome at a given utility point is not adjustable without changing the
perturbation budget globally.

\section{Conclusion}
\label{sec:conclusion}

This paper validates the \gls{fid}-constrained trajectory sharing mechanism in a
concrete \gls{isac} navigation setting. Simulation results on real pedestrian
traces confirm that a robot receiving privacy-controlled trajectory histories
can still perform reliable collision detection, with the information threshold
$\eta$ providing a single tunable parameter that jointly governs reconstruction
leakage and navigation utility. At the same missed-conflict rate,
\gls{fid}-controlled perturbation consistently achieves lower \gls{plr} and
shorter exposure lengths than fixed-noise schemes, because distortion is
concentrated on the most accurately sensed segments rather than applied
uniformly. These results support \gls{fid}-constrained sharing as a principled
data minimisation strategy for mobility-aware \gls{isac} deployments where
trajectory utility and personal privacy must be balanced simultaneously.
     \section*{Acknowledgment}
	This work is supported by the Federal Ministry of Research, Technology and Space of Germany via the project
Open6GHub+ (16KIS2406). B. Han (bin.han@rptu.de) is the
corresponding author.

\bibliographystyle{IEEEtran}
\bibliography{references}
\end{document}